\documentclass[runningheads,a4paper]{llncs}

\usepackage{amssymb}
\usepackage{graphicx}
\usepackage{url}
\usepackage{apacite}
\newcommand{\keywords}[1]{\par\addvspace\baselineskip
\noindent\keywordname\enspace\ignorespaces#1}
\usepackage{multirow}
\usepackage{hhline}
\begin{document}

\mainmatter  

\title{Signal-domain representation of symbolic music for learning embedding spaces}

%
%
\author{Mathieu Prang\inst{1} \and Philippe Esling\inst{1} \thanks{This work was supported by Jean-François Ducher through the construction of the principled synthetic dataset.}
}
%
\authorrunning{Mathieu Prang and Philippe Esling}
\titlerunning{{\em Signal-domain representation of symbolic music}, Full Paper}

\institute{IRCAM\\ 
\email{prang@ircam.fr, esling@ircam.fr}} 

%
%

\maketitle

\begin{abstract}
A key aspect of machine learning models lies in their ability to learn efficient intermediate features. However, the input representation plays a crucial role in this process, and polyphonic musical scores remain a particularly complex type of information.
In this paper, we introduce a novel representation of symbolic music data, which transforms a polyphonic score into a continuous signal. We evaluate the ability to learn meaningful features from this representation from a musical point of view. Hence, we introduce an evaluation method relying on principled generation of synthetic data. Finally, to test our proposed representation we conduct an extensive benchmark against recent polyphonic symbolic representations. 
We show that our signal-like representation leads to better reconstruction and disentangled features. This improvement is reflected in the metric properties and in the generation ability of the space learned from our signal-like representation according to music theory properties.
\end{abstract}

\section{Introduction}\label{sec:introduction}

Recent advances in machine learning and deep neural networks have produced extremely promising results. Notably, the fields of computer vision and Natural Language Processing (NLP) have been largely reshaped by these novel approaches \cite{mikolov2013efficient,bengio2009learning}. Based on these ideas, several models have also been proposed to enhance musical data analysis \cite{park2015music}. One of the major reasons behind these successes was the development of efficient embedding spaces for symbolic data \cite{mikolov2013distributed, pennington2014glove}. As these spaces encode semantic relationships between input data as metric properties, they allow to decompose a complex task into two separate and simpler sub-problems. First, a system learns an embedding that disentangles underlying semantic properties of a given type of data, which provides a continuous representation space. Then, another model can learn more advanced tasks based on this adequately organized space. Multiple attempts to mimic NLP embeddings for music have been proposed, based on the premises that they both target symbolic data \cite{huang2016chordripple,madjiheurem2016chord2vec,bretan2017learning}. However, these attempts only achieved small enhancements for musical generation and analysis tasks. This lack of adequate embeddings for symbolic music stems from the large discrepancies between the properties of musical and textual data. Indeed, the music vocabulary is smaller, highly repeated, and occurs in a wider variety of contexts in which the semantic properties of each element will be very different. Hence, it is not relevant to encode musical symbols as co-activation matrices \cite{pennington2014glove} or one-hot vectors \cite{mikolov2013distributed} as in the NLP field.

Recently, the success of Variational Auto-Encoder (VAEs) in symbolic music have been demonstrated in the MusicVAE model \cite{roberts2018hierarchical} which has provided interesting generation results. In addition to a new recurrent architecture, the novelty of this approach comes from the use of an efficient representation of input melodies. Nevertheless, as most recent proposals, this representation only takes into account monophonic melodies, which drastically reduces the scope of its application. Indeed, the musical composition process usually features intricate polyphony, rather than simply stacking monophonic voices together.
Motivated by this challenge, we propose in this paper a novel method, which allows to represent any symbolic score as a minimal audio signal. Different MIDI pitches are mapped to prime frequencies and summed across time resulting in a simplified waveform. The process being perfectly invertible, we can retrieve the original score without losing any information. We show that this novel representation (named \textit{signal-like}) is able to outperform previous propositions on polyphonic data, by comparing it to previous symbolic music representations, namely, the \textit{piano-roll}, the \textit{MIDI-like} \cite{oore2018time} and the \textit{NoteTuple} representation \cite{hawthorne2018transformer}. To do so, we compute the correlation between the learned embedding and music theory properties, allowing to quantify how different spaces are organized along musical concepts.

\section{State-of-the-art}\label{sec:stateofart}

\subsection{Embedding spaces}\label{subsec:wordembed}
The concept of embedding spaces has originally emerged in the NLP field. Multiple word embedding algorithms have been proposed, the two most notable and widely used contributions being Word2Vec \cite{mikolov2013efficient} and GloVe \cite{pennington2014glove}. 
By using such vectors as input representation for other machine learning tasks, NLP made colossal improvements. This success of embeddings has paved the way for a wide variety of applications such as sentiment analysis \cite{tang2014learning}, information retrieval \cite{palangi2016deep} or symbolic music field \cite{madjiheurem2016chord2vec, bretan2017learning, huang2016chordripple}.

More recently, the Variational Auto-Encoders (VAEs) \cite{kingma2013auto} have provided an elegant approach for learning embedding spaces. Similarly to a classical auto-encoder, these networks are composed of an \textit{encoder} $e(\mathbf{x}):\mathbb{R}^{x}\rightarrow\mathbb{R}^{z} $ which embeds the input in a lower dimensional \textit{latent} space, $\mathbb{R}^{z}$, followed by a \textit{decoder} $d(\mathbf{z}):\mathbb{R}^{z}\rightarrow\mathbb{R}^{x}$, which tries to reconstruct the original input, so that 
$$\hat{\mathbf{x}}=d(e(\mathbf{x}))\approx\mathbf{x}$$ 
Then, the generated outputs are compared to the inputs through a loss function that the network tries to minimize. In the case of VAEs, this loss function is defined by two different terms as follows
\begin{equation}
    \mathcal{L} \left(\mathbf{x}\mid\theta, \phi\right)= \mathbb{E}_{q_\theta(\mathbf{z}|\mathbf{x})}\left[\log p_\theta(\mathbf{x}|\mathbf{z})\right]+\mathcal{KL}\left(q_\theta(\mathbf{z}|\mathbf{x})||p(\mathbf{z})\right)
\end{equation}
The first term (expected log-likelihood) defines a \textit{reconstruction loss}, which pushes the decoder to produce outputs that are as close as possible to the inputs. Thus, it encourages learning an accurate reconstruction of the data. The second term (Kullback-Leibler divergence) acts as a \textit{regularizer}, enforcing information sharing between sample-wise distributions. Indeed, forcing the latent distribution of the data to be close to the normal distribution (with zero mean and unit variance) prevents the encoding network to isolate each projection, hence favoring close embedding vectors for similar inputs.

One of the most efficient models for symbolic music based on variational auto-encoding is \textit{MusicVAE} \cite{roberts2018hierarchical}. In this work, input scores are sliced in few bars and represented using a simplified version of the MIDI-like representation (see next section). 
The model is defined as a recurrent VAE, where the encoder is a two-layered bidirectional LSTM network that produces a sequence of hidden states $h=\{h_1,h_2,...,h_T\}$ from an input sequence $x=\{x_1,x_2,...,x_T\}$. The final encoding $z$ is obtained as a function of the last hidden state $h_T$. For the decoder, the authors rely on a hierarchical recurrent structure composed of two LSTM networks. The first one, called \textit{conductor} RNN, segments the output target into $U$ non-overlapping sub-sequences and produces an embedding vector $c=\{c_1,c_2,...,c_U\}$ for each time step. The second LSTM network auto-regressively produces a sequence of distributions over output tokens for each sub-sequence via a softmax output layer.

\subsection{Symbolic music representations}\label{subsec:symbolicrep}

\begin{figure}
 \includegraphics[width=1.02\textwidth]{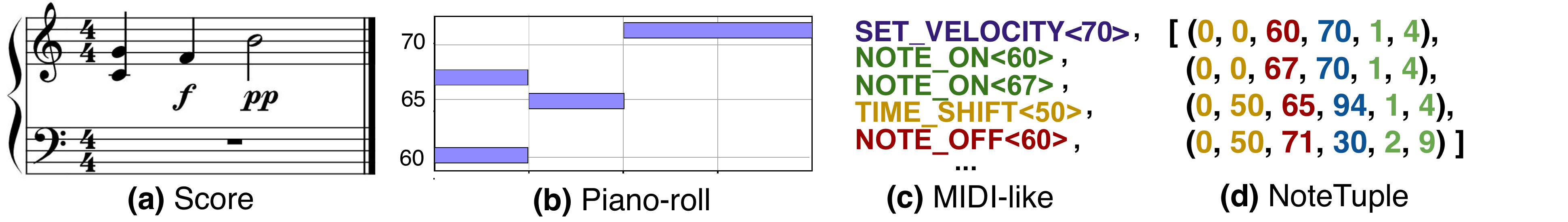}
 \caption{Representations for symbolic music learning from scores (a): the \textit{piano-roll} (b), the \textit{MIDI-like} representation (c) and the \textit{NoteTuple} representation (d).}
 \label{fig:midirep}
\end{figure}

The performances of machine learning techniques for symbolic generation is critically influenced by the properties of the input representation.
The most common way to represent polyphonic music is through the \textit{piano-roll} representation. Here, time is discretized with a reference quantum to provide a matrix $P(n,t)$ that represents note activation in musical sequences. Due to its sparsity and its repetitive nature, this representation poses several issues for learning, which warranted the definition of alternate approaches.

The first alternative representation has been proposed in \cite{oore2018time}. This \textit{MIDI-like} approach relies on an event-based vocabulary composed by four main MIDI events. The \texttt{NOTE\_ON} events with the corresponding \texttt{NOTE\_OFF} events, the \texttt{TIME\_SHIFT} events and the \texttt{SET\_VELOCITY} events. The resulting representation of an input piece is a variable-length sequence of discrete events taken from this vocabulary. This representation can handle any form of music with polyphony and variable number of voices or time signatures. An example is depicted in figure~\ref{fig:midirep} (c).

However, as the MIDI-like representation relies on the idea of time shifts, all the attributes corresponding to a given note (\textit{velocity}, \textit{note ON} and \textit{note OFF}) may be encoded at very distant positions of a sequence. To alleviate this particular issue, the \textit{NoteTuple} representation \cite{hawthorne2018transformer} was recently proposed. In this method, each note is represented by a tuple composed by four attributes, namely, the \textit{time offset} from the previous note, \textit{pitch}, \textit{velocity} and \textit{duration}. An example is displayed in figure~\ref{fig:midirep} (d).

\section{Signal-like representation}\label{sec:signalrep}

\begin{figure}
 \centerline{\includegraphics[width=0.85\columnwidth]{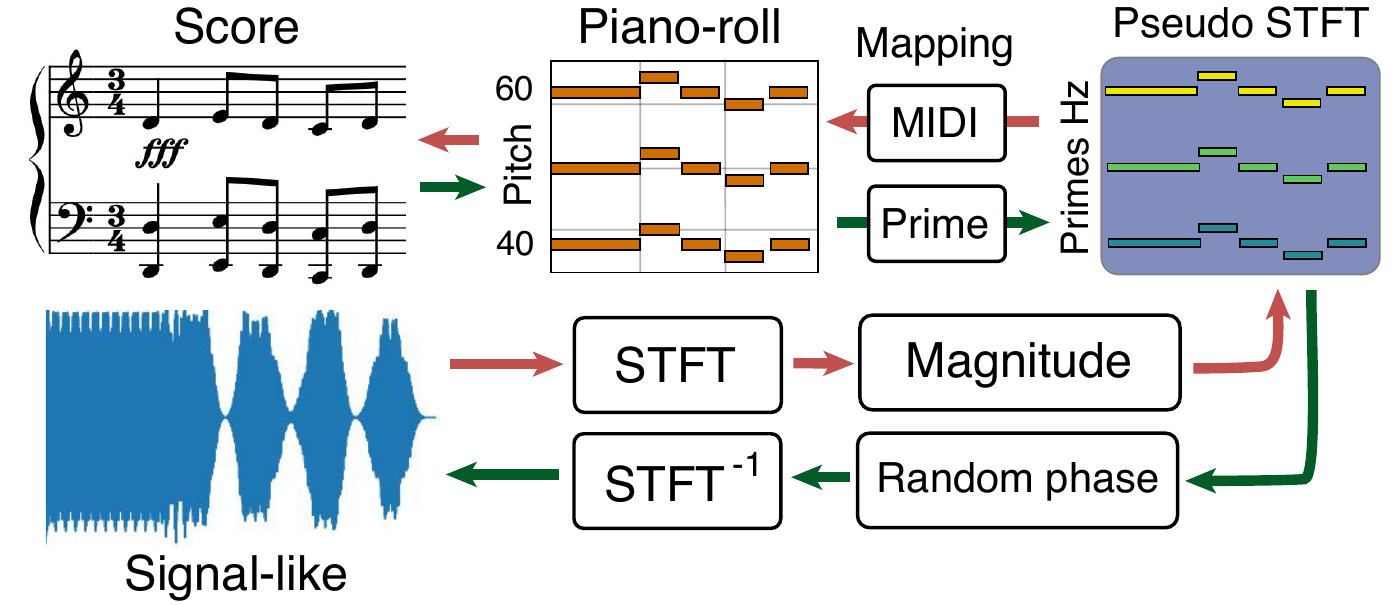}}
 \caption{Our proposed \textit{signal-like} representation. Based on a piano-roll matrix, each MIDI is mapped to prime frequencies, and an imaginary part equal to $1$ is added to each value of the matrix. Computing the inverse STFT on the resulting matrix produces the signal-like representation}
 \label{fig:sigrep}
\end{figure}


In machine learning applied to music, the audio signal information offers several desirable properties. Indeed, it naturally contains polyphonic information as a decomposable sum. Moreover, there has been some large successes in using raw signal for learning music generation tasks \cite{oord2016wavenet}. In this paper, we show that relying on a compact signal-like representation for polyphonic symbolic music can lead to large enhancements in tasks related to learning embedding. Hence, we aim to transform any given piano-roll as a spectrogram generating the most compact signal representation possible. To do so, we first map each MIDI pitch to prime numbers, starting from 43 (for MIDI pitch 0) to 2063 (for MIDI pitch 127) and removing consecutive numbers with a gap smaller than 3 (in order to not obtain pseudo frequencies that are too close together). In this representation, harmonic relationships may create phase effects detrimental to the inversion. Indeed, as the frequencies are arbitrary chosen and each represent a single MIDI pitch, the emergence of new harmonics during the inversion process would lead to the addition of notes which are not present in the original score. The use of prime numbers allow to mitigate this undesirable effect. Note that this also restrains the maximum frequency to a rather small value, allowing for a compact resulting signal. Then, we add an artificial phase, by setting imaginary parts to $1$ that create a complex matrix. Doing so, we can apply the inverse \textit{Short-Time Fourier Transform} (STFT) to the resulting matrix and obtain a compact signal-like representation of the score. In our experiments, we use a window size of 2048 with an hop size equal to $1$. Note that this whole process is invertible, as depicted in Figure
~\ref{fig:sigrep}.

\section{Evaluation}\label{sec:eval}

In order to precisely evaluate how embedding spaces are able to capture musical theory principles, we design a principled generative approach. This allows to obtain large sets of evaluation data with controlled properties, while following known music theory rules. For our experiments, we produce a dataset composed by bars of synthetic chorales (see the support page for details).\footnote{\url{https://github.com/MagiCzOOz/signallike-embedding}} We obtain a set of 21966 realisations from 370 different skeletons where the links between each realisation and its corresponding skeleton have been kept. In this paper, this corpus is used only for evaluation purpose. Hence, none of these data are used during the training. 

Based on this corpus, we can analyze different aspects of the organization of learned embedding spaces. First, we can analyze the behavior of different spaces based on tonality. An adequate clustering of the different tonalities across the space would prove its capacity to encode this high level musical information. Second, we can compute the distance between a realisation and its corresponding skeleton based on the number of non-harmonic tones. In this case, if the space is well-organized, we expect these distances to evolve linearly with the number of additional tones. Third, we can compute the distance between a realisation and its corresponding skeleton based on the type of non-harmonic tones. A link between these distances and the note type would show that the embedding space can handle advanced musical concepts.

\section{Experiments}\label{sec:experiments}

In order to compare the efficiency of \textit{piano-roll}, \textit{MIDI-like}, \textit{NoteTuple} and \textit{Signal-like} representations, we evaluate them with the same common learning approach. 
Based on the success of the MusicVAE model \cite{roberts2018hierarchical}, we rely on this architecture for our experiments. We use a recurrent encoder with a two-layers bidirectional LSTM of 1024 units. For the decoder, we rely on a hierarchical recurrent architecture with a two-layer unidirectional LSTM with a hidden size of 1024 for both the conductor and decoder. We use a latent size of 256 which largely reduces the input dimensionality while keeping enough information for the reconstruction.

As discussed previously, we train the models on the JSB chorales dataset. We use data augmentation based on transposition as proposed in \cite{hadjeres2017deepbach}, ensuring that transposed chorales are still correct from a music theory standpoint. This leads to an extended set of 2418 training and 549 testing chorales. As the model takes musical bars as input, we split the MIDI files into separate bars and compute different representations on each bar. We filter bars to have a maximum of 64 events, retaining more than $95\%$ of all the data. After these operations, we obtain 36801 training and 8850 testing bars. Regarding the MIDI-like representation, we introduce a dummy event as padding to obtain a constant-size representation. We use a similar approach for the NoteTuple representation by taking 16 tuples by bar and padding with empty tuples.

All the models are trained with the ADAM optimizer \cite{kingma2014adam} with an initial learning rate of $10^{-4}$ and a batch size of $16$. In order to train the \textit{piano-roll} and \textit{signal-like} representation, we rely on a Mean Squared Error (MSE) loss. Since the MIDI-like and NoteTuple representation are categorical, we rather rely on a cross-entropy loss.

\section{Results}\label{sec:results}

\subsection{Reconstruction}

First, we compare the capacity of each model to reconstruct the original data, by relying on a frame-level accuracy measure. We divide musical sequences into 16 frames and compute the accuracy as a ratio involving correct (true positives) and wrong (false negatives and positives) active notes in each one of them.
We also display separately the best reconstruction accuracy and KL divergence results on the test set in Table~\ref{tab:tablres}.

\begin{table}
\small
\begin{center}
\caption{Reconstruction accuracy and KL divergence on the test set for different input representations.} \label{tab:tablres}
\begin{tabular}{l|c|c|c|}
  \cline{2-4}
   & Input & \begin{tabular}{@{}c@{}}Reconstruction \\ accuracy (\%)\end{tabular} & KL div \\
  \hline
  \multirow{2}{*}{Monophonic} & Piano-roll & $95.8$ & $2*10^3$ \\
  & MIDI-like-mono & $97.5$ & $1*10^3$ \\
  \hhline{=|=|=|=|}
  \multirow{4}{*}{Polyphonic} & Piano-roll & $94.1$ & $2*10^3$   \\ 
  & MIDI-like & $ <1 $ & -  \\
  & NoteTuple & $17.3$ & $9*10^4$ \\
  & Signal-like & \textbf{96.5} & $ \mathbf{1*10^3} $  \\
  \hline
\end{tabular}
\vspace{-0.5cm}
\end{center}
\end{table}

As we can see, the \textit{MIDI-like} shows great performance in the reconstruction of monophonic musical bars but are unable to achieve the reconstruction with polyphonic ones. Indeed, due to the nature of the \textit{MIDI-like} representation, even a unique error on a \texttt{NOTE\_ON} or \texttt{NOTE\_OFF} event leads to an ill-defined musical sequences with notes that never end or never start. Concerning \textit{NoteTuple}, the models have been able to encode information about the number of notes, duration and time offsets, which are almost perfectly reconstructed. However, a large number of mistakes in the pitches of individual notes cause the frame-level accuracy score to be low. Hence, despite the use of regularization techniques (dropout, data augmentation) these two representation seem to largely suffer from over-fitting.

On the other hand, great results have been achieved with the \textit{signal-like} representation. In addition to a better reconstruction accuracy, our representation has improved learning stability by avoiding the exploding gradient problem \cite{pascanu2012understanding} which occurred with the piano-roll representation. Moreover, the robustness of our proposal decreases the loss of reconstruction information while minimizing the KL div and thus lead to a much better trade-off score.

\subsection{Music theory analysis}

\begin{figure}
 \centerline{\includegraphics[width=12.5cm]{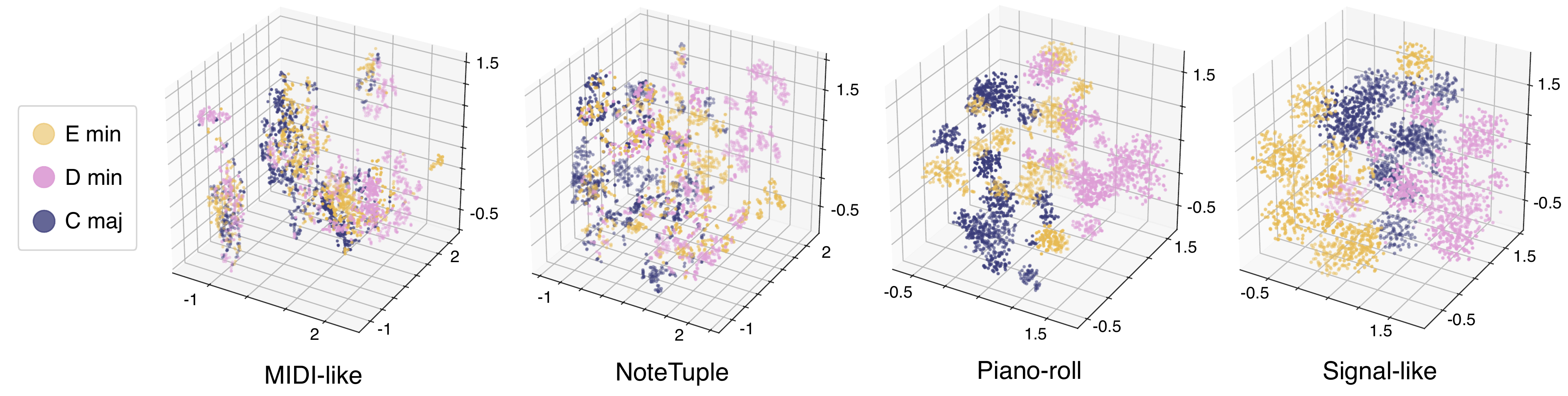}}
 \caption{t-SNE plots of different latent spaces with our unseen synthetic data. The colors represent the tonalities.}
 \label{fig:tonality}
\end{figure}

First, we embed unseen synthetic bars in order to compare their respective projections in the different embeddings. As the latent spaces have 128 dimensions, precluding direct visualization, we rely on a t-SNE dimensionality reduction \cite{maaten2008visualizing} with 1000 iterations and a perplexity of 30. We display in Figure~\ref{fig:tonality} the results of this analysis. It clearly appears that the use of our proposed \textit{signal-like} representation greatly improves the efficiency of the latent space in encoding a unseen high-level musical feature such as the tonality.

\begin{figure}
 \centerline{\includegraphics[width=0.45\columnwidth]{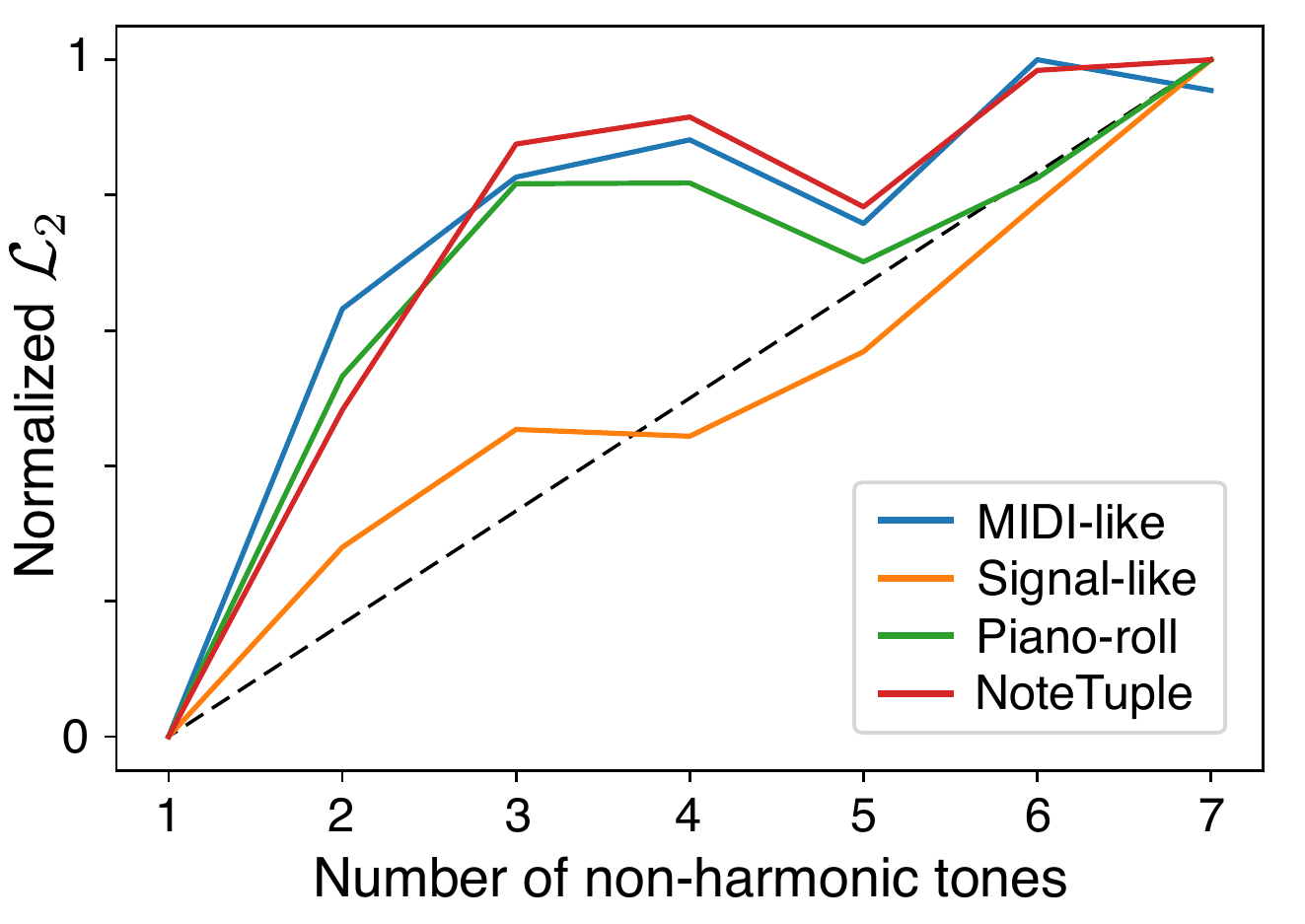}}
 \caption{Normalized mean $\mathcal{L}_{2}$ distances between the original skeleton and its realizations depending on the number of non-harmonic tones.}
 \label{fig:distance}
\end{figure}

Then, we compute statistics over metadata of the toy dataset in order to evaluate how different spaces have organized the realizations with respect to the number of non-harmonic tones. As we can see in Figure~\ref{fig:distance}, the distances between the original skeleton and its different realizations in the space learned through the signal-like representation provide an almost linear relationship with the number of non-harmonic tones. This highlights the fact that the \textit{signal-like} space is better organized from a musical point of view.

\vspace{-0.7cm}

\subsection{Symbolic music generation}
To highlight the generative capacity of the latent spaces trained by using our signal-like representation, we generate interpolations between two random points in the latent space, which are available on the support page.\footnote{\url{https://github.com/MagiCzOOz/signallike-embedding}} The generations generally provide smooth evolution from one point to the other, even though the two underlying musical sequences are largely different. Hence, this shows that our \textit{signal-like} representation can be used in creative applications, while providing organized latent structures.

\section{Conclusion}\label{sec:conclusion}

In this paper, we have proposed a new representation for symbolic music and evaluated it through the learning of musical embedding spaces. We have shown that our signal-like representation have improved the quality of the learned spaces for both the reconstruction quality and the organization of the underlying spaces.
In future work, we are interested in optimizing the model for our representation to understand various musical qualities. Furthermore, in addition to the pure random latent generation, several creative applications could be explored by relying on these symbolic embedding spaces.

\bibliographystyle{apacite}
\bibliography{CSMC_MUME_LaTeX_Template.bib}

\end{document}